\documentclass[runningheads]{llncs}
\usepackage[T2A]{fontenc}
\usepackage[english]{babel}
\usepackage{amsmath}
\usepackage{amssymb}
\usepackage{booktabs}
\usepackage{graphicx}
\usepackage{color}
\usepackage{subfigure}

\newcounter{todocnt}
\newcommand{\todo}[1][$\bullet\bullet\bullet$]{\refstepcounter{todocnt} \textcolor{blue}{$\bullet\bullet\bullet$\ To do (\thetodocnt): #1}}

\renewcommand{\todo}[1]{}

\begin{document}
\title{Q-Mixing Network for Multi-Agent Pathfinding in Partially Observable Grid Environments}
\titlerunning{QMIX in Milti-Agent Pathfinding}

\author{
    Vasilii Davydov\inst{2, 3} \and 
    Alexey Skrynnik\inst{1} \and
    Konstantin Yakovlev\inst{1, 2} \and
    Aleksandr~I.~Panov\inst{1, 2}
}
\authorrunning{V. Davydov and etc.}

\institute{
Artificial Intelligence Research Institute FRC CSC RAS, Moscow, Russia \and
Moscow Institute of Physics and Technology, Moscow, Russia \and
Moscow Aviation Institute, Moscow, Russia
}

%

\maketitle

\begin{abstract}
In this paper, we consider the problem of multi-agent navigation in partially observable grid environments. This problem is challenging for centralized planning approaches as they, typically, rely on the full knowledge of the environment. We suggest utilizing the reinforcement learning approach when the agents, first, learn the policies that map observations to actions and then follow these policies to reach their goals. To tackle the challenge associated with learning cooperative behavior, i.e. in many cases agents need to yield to each other to accomplish a mission, we use a mixing Q-network that complements learning individual policies. In the experimental evaluation, we show that such approach leads to plausible results and scales well to large number of agents.

\keywords{Multi-agent Pathfinding \and Reinforcement Learning \and Mixing Q-network \and Grid Environment}
\end{abstract}

\section{Introduction}
Planning the coordinated movement for a group of mobile agents is usually considered as a standalone problem within the behavior planning. Two classes of methods for solving this problem can be distinguished: centralized and decentralized. The methods of the first group are based on the assumption of the existence of a control center, which has access to complete information about the states and movements of agents at any time. In most cases, such methods are based either on reducing the multi-agent planning to other well-known problems, e.g. the boolean satisfiability (SAT-problems)~\cite{surynek2016efficient}, or on heuristic search. Among the latter, algorithms of the conflict-based search (CBS) family are actively developing nowadays. Original CBS algorithm~\cite{sharon2015conflict} guarantees completeness and optimality. Many enhancements of CBS exist that significantly improve its performance while preserving the theoretical guarantees -- ICBS~\cite{boyarski2015icbs}, CBSH~\cite{felner2018adding} etc. Other variants of CBS, such as the ones that take the kinematic constraints~\cite{andreychuk2020multi} into account, target bounded-suboptimal solutions~\cite{barer2014suboptimal} are also known. Another widespread approach to centralized multi-agent navigation is prioritized planning~\cite{yakovlev2019prioritized}. Prioritized planners are extremely fast in practice, but they are incomplete in general. However, when certain conditions have been met the guarantee that any problem will be solved by a prioritized planning method can be provided~\cite{cap2015a}. In practice these conditions are often met in the warehouse robotics domains, therefore, prioritized methods are actively used for logistics applications.

Methods of the second class, decentralized, assume that agents are controlled internally and their observations and/or communication capabilities are limited, e.g. they do not have direct access to the information regarding other agent's plans. These approaches are naturally better suited to the settings when only partial knowledge of the environment is available. In this work we focus on one such setting, i.e. we assume that each agent has a limited field of view and can observe only a limited myopic fragment of the environment.

Among the methods for decentralized navigation, one of the most widely used is the ORCA algorithm~\cite{van2011reciprocal} and its numerous variants. These algorithms at each time step compute the current speed via the construction of the so-called velocity obstacle space. When certain conditions are met, ORCA guarantees that the collision between the agents is avoided, however, there is no guarantee that each agent will reach its goal. In practice, when navigating in a confined space (e.g. indoor environments with narrow corridors, passages, etc.), agents often find themselves in a dead-lock, when they reduce their speed to zero to avoid collisions and stop moving towards goals. It is also important to note that the algorithms of the ORCA family assume that velocity profiles of the neighboring agents are known. In the presented work, such an assumption is not made and it is proposed to use learning methods, in particular -- reinforcement learning methods, to solve the considered problem.

Application of the reinforcement learning algorithms for path planning in partially observable environments is not new \cite{Shikunov2019,Martinson2020}. In~\cite{panov2018grid} the authors consider the single-agent case of a partially observable environment and apply the deep Q-network~\cite{mnih2015human} (DQN) algorithm to solve it.

In~\cite{sartoretti2019primal} the multi-agent setting is considered. The authors suggest using the neural network approximator that fits parameters using one of the classic deep reinforcement learning algorithms. However, the full-fledged operation of the algorithm is possible only when the agent's experience is complimented with the expert data generated by the state-of-the-art centralized multi-agent solvers. The approach does not use maximization of the general Q-function but tries to solve the problem of multi-agent interaction by introducing various heuristics: an additional loss function for blocking other agents; a reward function that takes into account the collision of the agents; encoding other agents' goals in the observation.

In this work we propose to use a mixing Q-network, which implements the principle of monotonic improvement of the overall assessment of the value of the current state based on the current assessments of the value of the state of individual agents. Learning the mixing mechanism based on a parameterized approximator allows to automatically generate rules for resolving conflict patterns when two or more agents pass intersecting path sections. We also propose the implementation of a flexible and efficient experimental environment with trajectory planning for a group of agents with limited observations. The configurability and the possibility of implementing various behavioral scenarios by changing the reward function allow us to compare the proposed method with both classical methods of multi-agent path planning and with reinforcement learning methods designed for training one agent.

\section{Problem statement}

We reduce the multi-agent pathfinding problem in partially observable environment to a single-agent pathfinding with dynamic obstacles (other agents), as we assume decentralized scenario (i.e. each agent is controlled separately). 
The process of interaction between the agent and the environment is modeled by the partially observable Markov decision process (POMDP), which is a variant of the (plain) Markov decision process. We sought a policy for this POMDP, i.e. a function that  maps the agent's observations to actions. To construct such a function we utilize reinforcement learning.

Markov decision process is described as the tuple $(S, A, P, r, \gamma)$. At every step the environment is assumed to be at the certain state $s \in S$ and the agent has to decide which action $a \in A$ to perform next. After picking and performing the action the agent receives a reward via the reward function $r(s, a): S \times A \rightarrow {\mathbb R}$. The environment is (stochastically) transitioned to the next state $P(s'|s, a): S \times A \times S \rightarrow [0, 1]$. The agent chooses its action based on the policy $\pi$ which is a function $\pi(a|s): A \times S \rightarrow [0, 1]$. In many cases, it is preferable to choose the action that maximizes the Q-function of the state-action pair: $Q(a_t, s_t) = r(s_t, a_t) + \mathbb E ( \sum_{i=1}^\infty \gamma^i r(s_{t+i}, a_{t+i} ) )$, where $\gamma<1$ is the given discount factor.

The POMDP differs from the described setting in that a state of the environment is not fully known to the agent. Instead it is able to receive only a partial observation $o \in O$ of the state. Thus POMDP is a tuple $(S, O, A, P, r, \gamma)$. The policy now is a mapping from observations to actions: $\pi(a|o): A \times O \rightarrow [0, 1]$. Q-function in this setting is also dependent on observation rather than on state: $Q(a_t, o_t) = r(s_t, a_t) + \mathbb E ( \sum_{i=1}^\infty \gamma^i r(s_{t+i}, a_{t+i} ) )$.

In this work we study the case when the environment is modeled as a grid composed of the free and blocked cells. The actions available for an agent are: move left/up/down/right (to a neighboring cell), stay in the current cell. We assume that transitions are non-stochastic, i.e. each action is executed perfectly and no outer disturbances that affect the transition are present. The task of an agent is to construct a policy that picks actions to reach the agent's goal while avoiding collisions with both static and dynamic obstacles (other agents). In this problem, the state $s \in S$ describes the locations of \emph{all} obstacles, agents, and goals. Observation $o \in O$ contains the information about the location of obstacles, agents, and goals \emph{in the vicinity} of the agent (thus, the agent sees only its nearby surrounding).

\section{Method}

In this work, we propose an original architecture for decision-making and agent training based on a mixing Q-network that uses deep neural network to parameterize the value function by analogy with deep Q-learning (DQN).

In deep Q-learning, the parameters of the neural network are optimized $Q(a, s| \theta )$. Parameters are updated for mini-batches of the agent's experience data, consisting of sets $<s, a, r, s'>$, where $s'$ is the state in which the agent moved after executing action $a$ in the state $s$.

The loss function for the approximator is:

\[ L=\sum_{i=1}^{b} [((r^i+\gamma max_{a^{i+1}}Q(s^{i+1}, a^{i+1}|\theta)) - Q(s^i, a^i |\theta))^2], \] 
b is a batch size.

In the transition to multi-agent reinforcement learning, one of the options for implementing the learning process is independent execution. In this approach, agents optimize their own Q-functions for the actions of a single agent. This approach differs from DQN in the process of updating the parameters of the approximator when agents use the information received from other agents. In fact, agents decompose the value function (VDN)~\cite{Sunehag2017} and aim to maximize the total Q-function $Q_{tot} (\tau, u)$, which is the sum of the Q-functions of each individual agent $Q^i (a^i, s^i | \theta^i)$.

The Mixing Q-Network (QMIX)~\cite{Rashid2018} algorithm works similarly to VDN. However, in this case, to calculate $Q_{tot}$, a new parameterized function of all Q-values of agents is used. More precisely, $Q_{tot}$ is calculated to satisfy the condition that $Q_{tot}$ increases monotonically with increasing $Q^i$:

\[ \frac{\delta Q_{tot} }{\delta Q^i} \geq 0 \; \forall i \; 1 \leq i \leq num \, agents, \]
$Q_{tot}$ is parameterized using a so-called mixing neural Q-network. The weights for this network are generated using the hyper networks~\cite{Ha2016}. Each of the hyper networks consists of a single fully connected layer with an absolute activation function that guarantees non-negative values in the weights of the mixing network. Biases for the mixing network are generated similarly, however they can be negative. The final bias of the mixing network is generated using a two-layer hyper network with ReLU activation.

\begin{figure}
\centerline{\includegraphics[ width=0.8\linewidth]{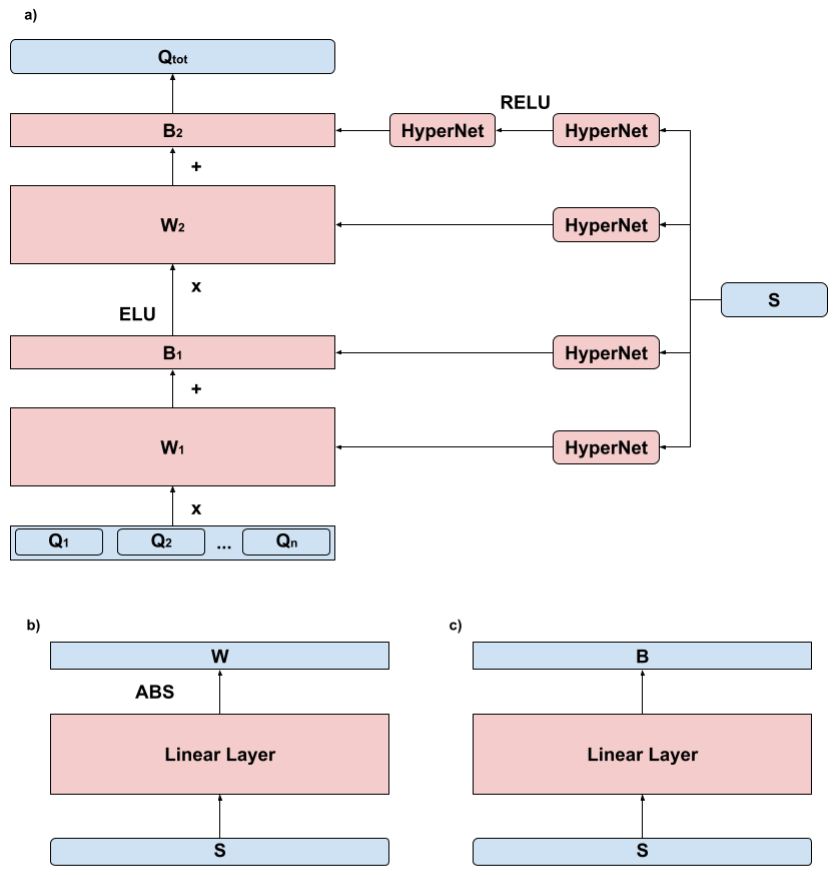}}
    \caption{a) Mixing network architecture. $W_1$, $W_2$, $B_1$, $B_2$ are the weights of the mixing network; $Q_1$, $Q_2$ ... $Q_n$ are the agents' Q values; $s$ is the environment state; $Q_{tot}$ is a common Q Value; b) Hyper network architecture for generating the weights matrix of the mixing Q-network. The hyper network consists of a single fully connected layer and an absolute activation function. c) Hyper network architecture for generating the biases of the mixing Q-network. The hyper network consists of a single fully connected layer.}
\label{fig:qmix-scheme}
\end{figure}

The peculiarities of the mixing network operation also include the fact that the agent's state or observation is not fed into the network, since $Q_{tot}$ is not obliged to increase monotonically when changing the state parameters $s$. Individual functions $Q^i$ receive only observation as input, which can only partially reflect the general state of the environment. Since the state can contain useful information for training, it must be used when calculating $Q_{tot}$, so the state $s$ is fed as the input of the hyper network. Thus, $Q_{tot}$ indirectly depends on the state of the environment and combines all Q-functions of agents. The mixing network architecture is shown in Figure~\ref{fig:qmix-scheme}.

The mixing network loss function looks similar to the loss function for DQN:

\[ \sum_{i=1}^{b} [({y^i_{tot}} - Q_{tot}(\tau^i, u^i, s^i|\theta))^2] \] 

\[ y_{tot}=r+\gamma max_{t^{i+1}}Q_{tot}(\tau^{i+1}, u^{i+1}, s^{i+1}|\theta^-). \]
Here $b$ is the batch size, $\tau^{i+1}$ is the action to be performed at the next step after receiving the reward $r$, $u^{i+1}$ is the observation obtained at the next step, $s^{i+1}$ is the state obtained in the next step. $\theta^-$ are the parameters of the copy of the mixing Q-network created to stabilize the target variable.

\section{Experimental environment for multi-agent pathfinding}
\begin{figure*}[ht]
    \centering
    \subfigure{\includegraphics[width=0.42\linewidth]{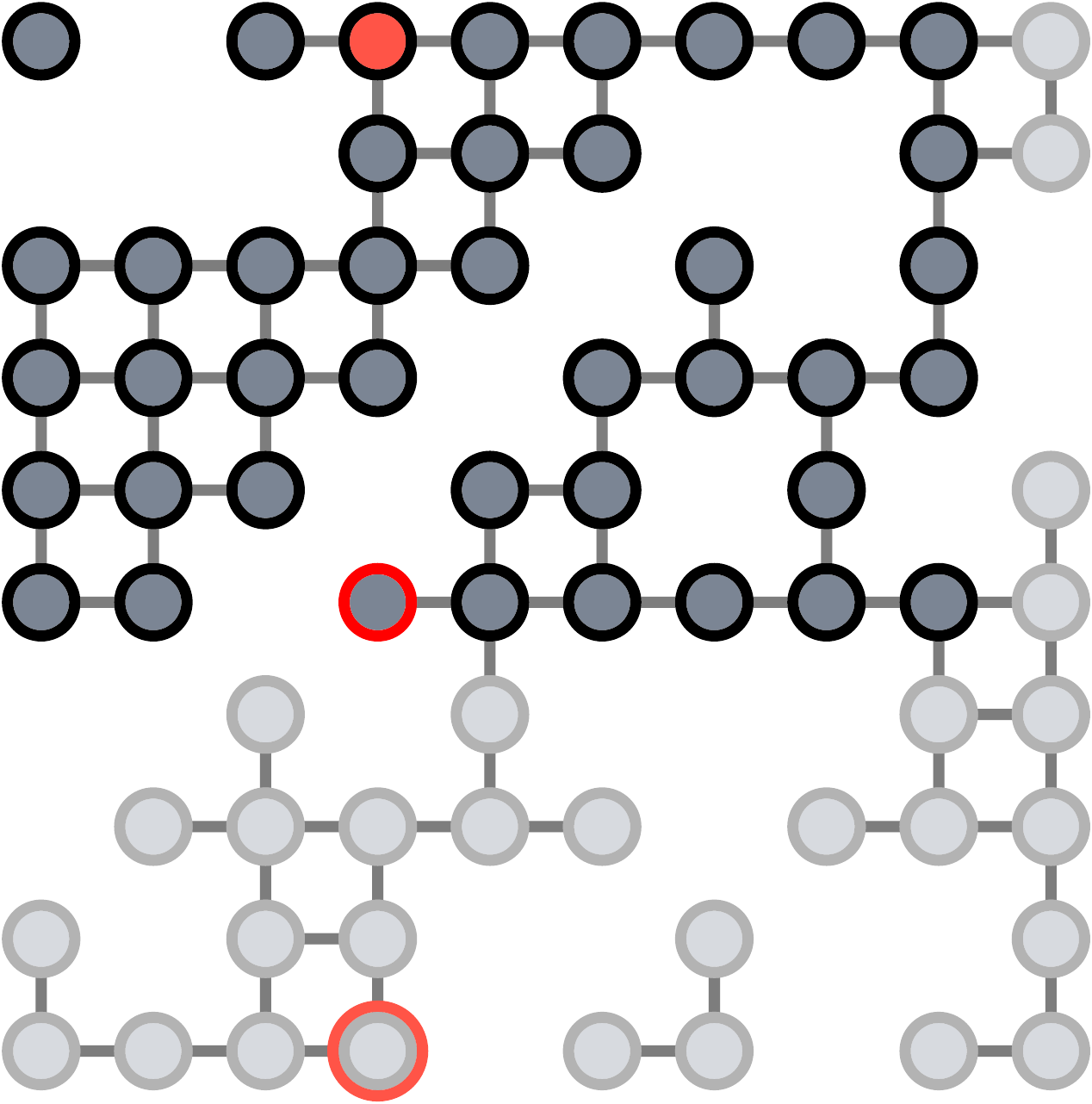}}
    \hspace{0.1\linewidth}
    \subfigure{\includegraphics[width=0.42\linewidth]{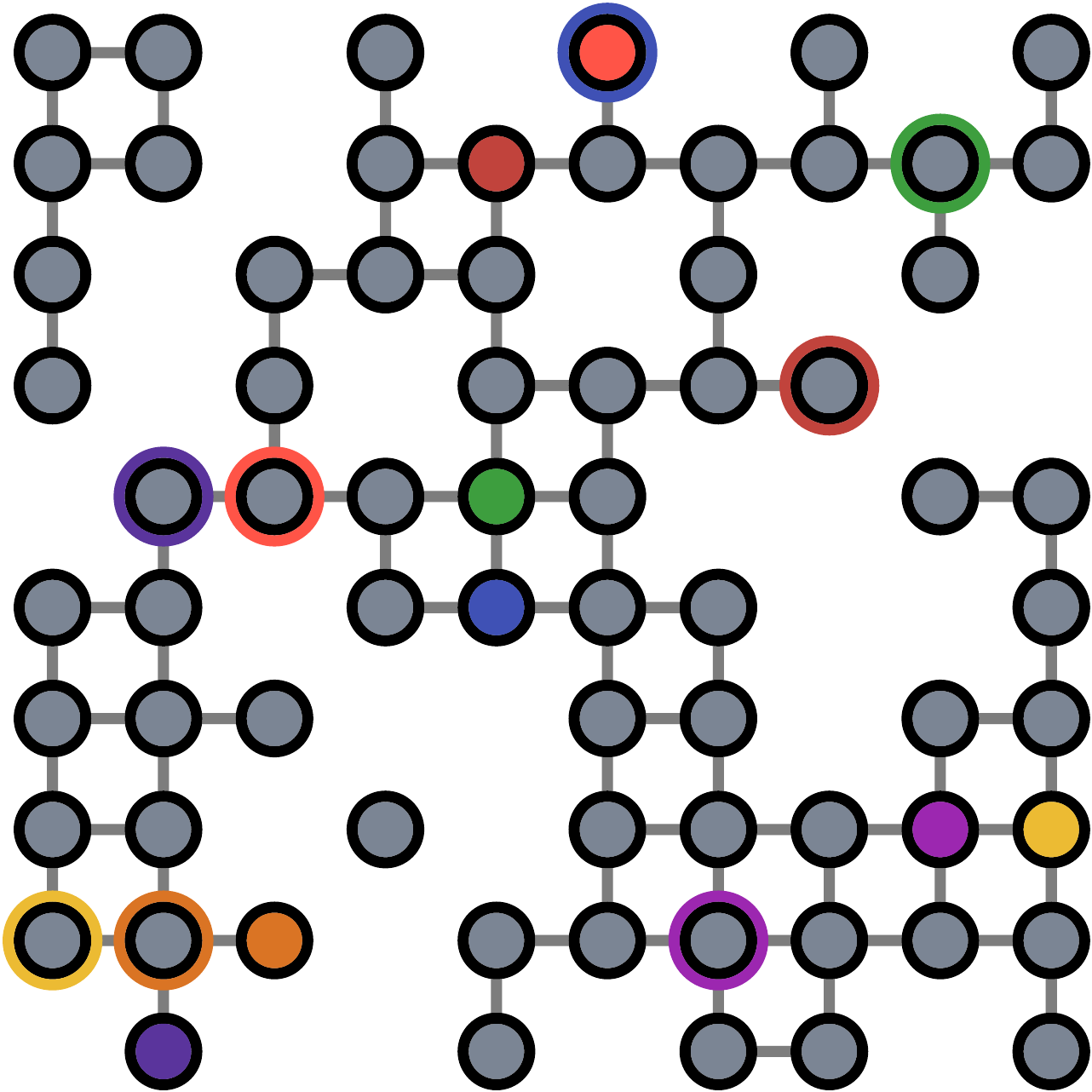}}
\caption{The left figure shows an example of partial observability for a single agent environment: gray vertices are free cells along the edges of which the agent can move; a filled red circle indicates the position of the agent; the vertex with a red outline is the target of this agent, vertex with red border - projection of the agent's goal. The area that the agent cannot see is shown as transparent. The right figure shows an example of an environment for 8 agents, projections of agents' goals and partial observability are not shown for visibility purposes.}
    \label{fig:env-example}
\end{figure*}

The environment is a grid with agents, their goals, and obstacles located on it. Each agent needs to get to his goal, avoiding obstacles and other agents. An example of an environment showing partial observability for a single agent is shown in Figure~\ref{fig:env-example}. This figure also shows an example of a multi-agent environment.

The input parameters for generating the environment are: 
\begin{itemize}
    \item field size $E_{size} \geq 2$,
    \item obstacle density $E_{density}\in[0, 1)$,
    \item number of agents in the environment $E_{agents} \geq 1$,
    \item observation radius: agents get $1 \leq R \leq E_{size}$ cells in each direction,
    \item the maximum number of steps in the environment before ending $E_{horizon} \geq 1$,
    \item the distance to the goal for each agent $E_{dist}$ (is an optional parameter, if it is not set, the distance to the goal for each agent is generated randomly). 
\end{itemize}

Obstacle matrix is filled randomly by parameters $E_{size}$ and $E_{density}$. The positions of agents and their goals are also generated randomly, but with a guarantee of reachability.

\begin{figure}
\centerline{\includegraphics[width=\textwidth]{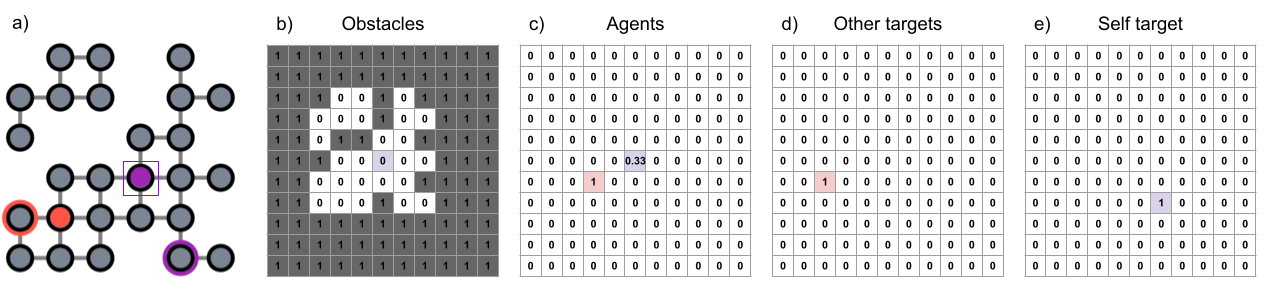}}
\caption{An example of an observation matrix for a purple agent. In all observation cells, 1 means that there is an object (obstacle, agent, or goal) in this cell and 0 otherwise. a) Environment state. The agent for which the observation is shown is highlighted; b) Obstacle map. The central cell corresponds to the position of the agent, in this map the objects are obstacles; c) Agents map. In this map, the objects are agents; d) Other agents' goals map. In this map, the objects are the goals of other agents; e) Goal map. In this map, the object is the self-goal of the agent.}
\label{fig4}
\end{figure}

The observation space $O$ of each agent is a multidimensional matrix: $O: 4 \times \big(2\times R + 1\big) \times \big(2\times R + 1 \big)$, which includes the following 4 matrices. \textit{Obstacle matrix}: $1$ encodes an obstacle, and $0$ encodes its absence. If any cell of the agent's field of view is outside the environment, then it is encoded $1$. \textit{Agents' positions matrix}: $1$ encodes other agent in the cell, and $0$ encodes his absence. The value in the center is inversely proportional to the distance to the agent's goal. \textit{Other agents' goals matrix}: $1$ if there is a goal of any agent in the cell, $0$ -- otherwise. \textit{Self agent's goal matrix} if the goal is inside the observation field, then there is $1$ in the cell, where it is located, and $0$ in other cells. If the target does not fall into the field of view, then it is projected onto the nearest cell of the observation field. As a cell for projection, a cell is selected on the border of the visibility area, which either has the same coordinate along with one of the axes as the target cell or if there are no such cells, then the nearest corner cell of the visibility area is selected. An example of an agent observation space is shown in Figure~\ref{fig4}.

Each agent has 5 actions available: stay in place and move vertically (up or down) or horizontally (right or left). An agent can move to any free cell that is not occupied by an obstacle or other agent. If an agent moves to a cell with his own goal, then he is removed from the map and the episode is over for him.

Agents receive a reward of $0.5$ if he follows one of the optimal routes to his goal, $-1$ if the agent has increased his distance to the target and $-0.5$ if the agent stays in place.

\section{Experiments}
This section compares QMIX with the Proximal Policy Optimization (PPO), single-agent reinforcement learning algorithm~\cite{schulman2017proximal}. We chose PPO because it showed better results in a multi-agent setting compared to other modern reinforcement learning algorithms. Also, this algorithm significantly outperformed other algorithms, including QMIX, in the single-agent case.

The algorithms were trained on environments with the following parameters: $ E_{size} = 15 \times15 , E_{density} = 0.3 $, $ E_{agents} = 2 $, $ R = 5 $, $ E_{horizon} = 30 $, $ E_{dist} = 8 $. As the main network architecture of each of the algorithms, we used architecture with two hidden layers of $ 64 $ neurons, with  ReLU activation function for QMIX and Tanh for PPO. We trained each of the algorithms using 1.5M steps in the environment.

\begin{figure*}[ht]
    \centering
    \subfigure{\includegraphics[width=0.49\linewidth]{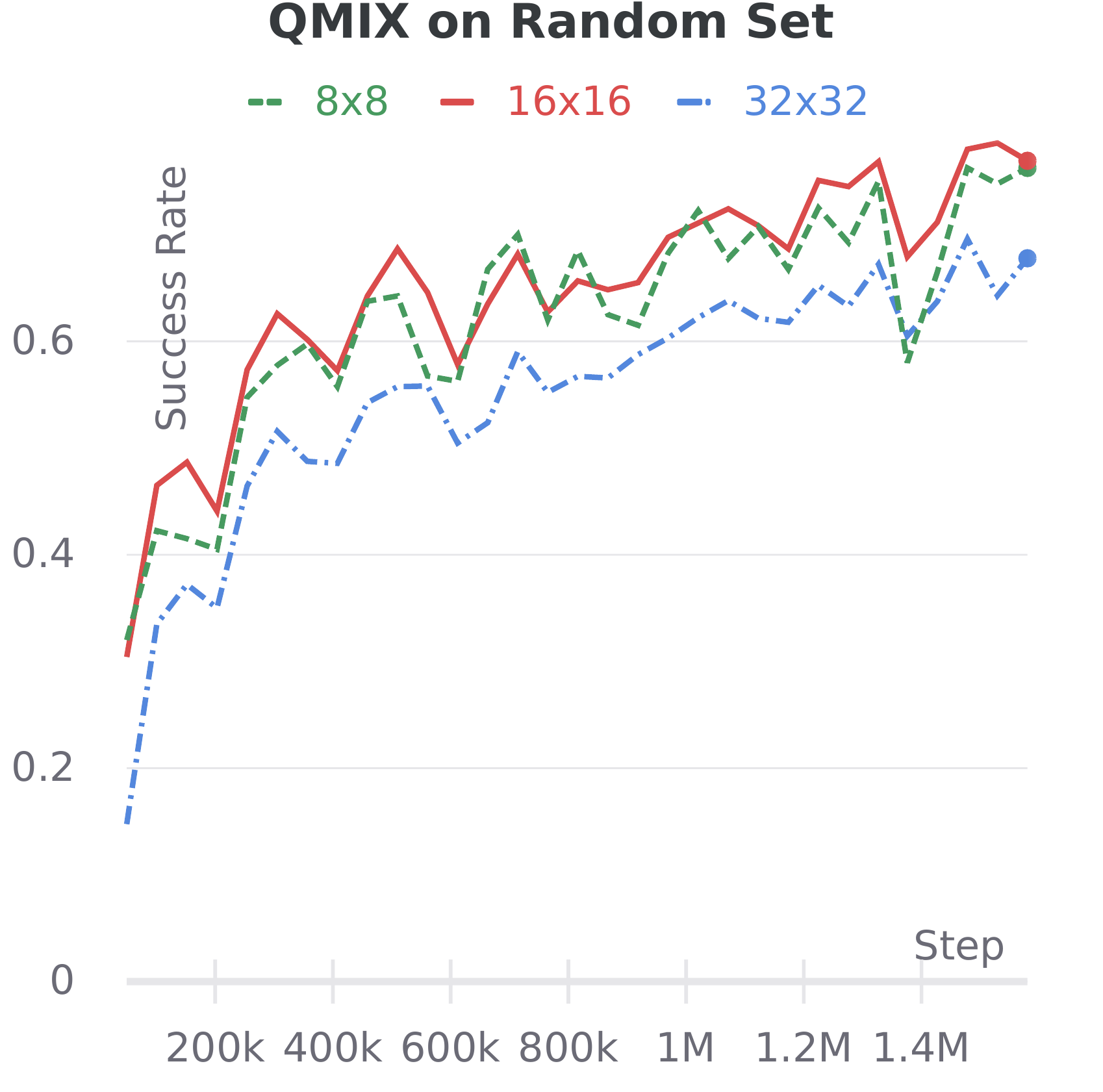}}
    \subfigure{\includegraphics[width=0.49\linewidth]{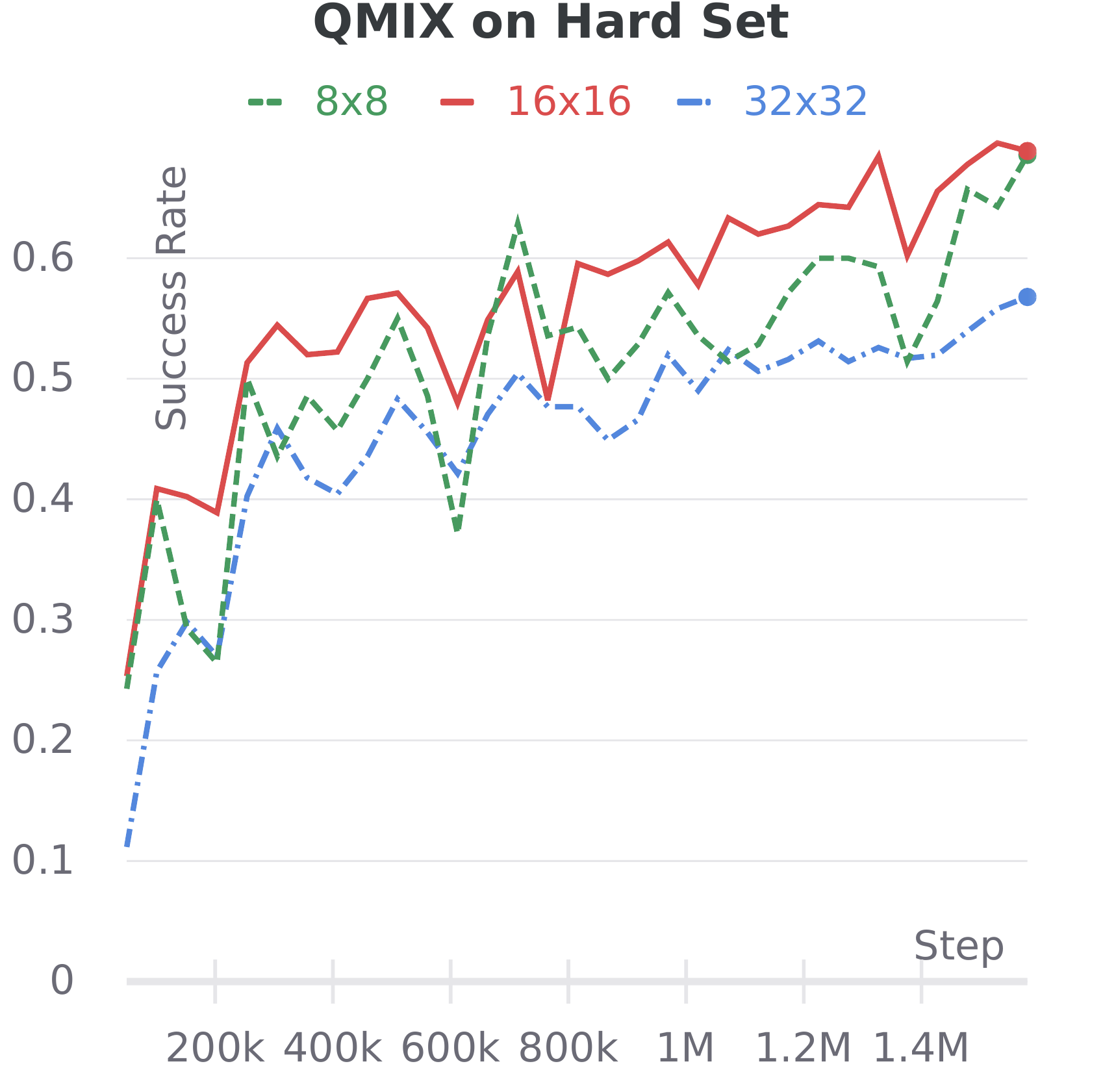}}

    \caption{The graphs show separate curves for different environment sizes. 
    For environment sizes $ 8 \times8 $; $ 16 \times16 $; $ 32 \times32 $, we used 2, 6, 16 agents, respectively.
    The left graph shows the success rate for the QMIX algorithm in random environments. The right graph shows the success rate for the QMIX algorithm in complex environments.}
    \label{fig:qmix-training}
\end{figure*}

\begin{figure*}[htp]
    \centering
    
    \subfigure{\includegraphics[width=0.29\linewidth]{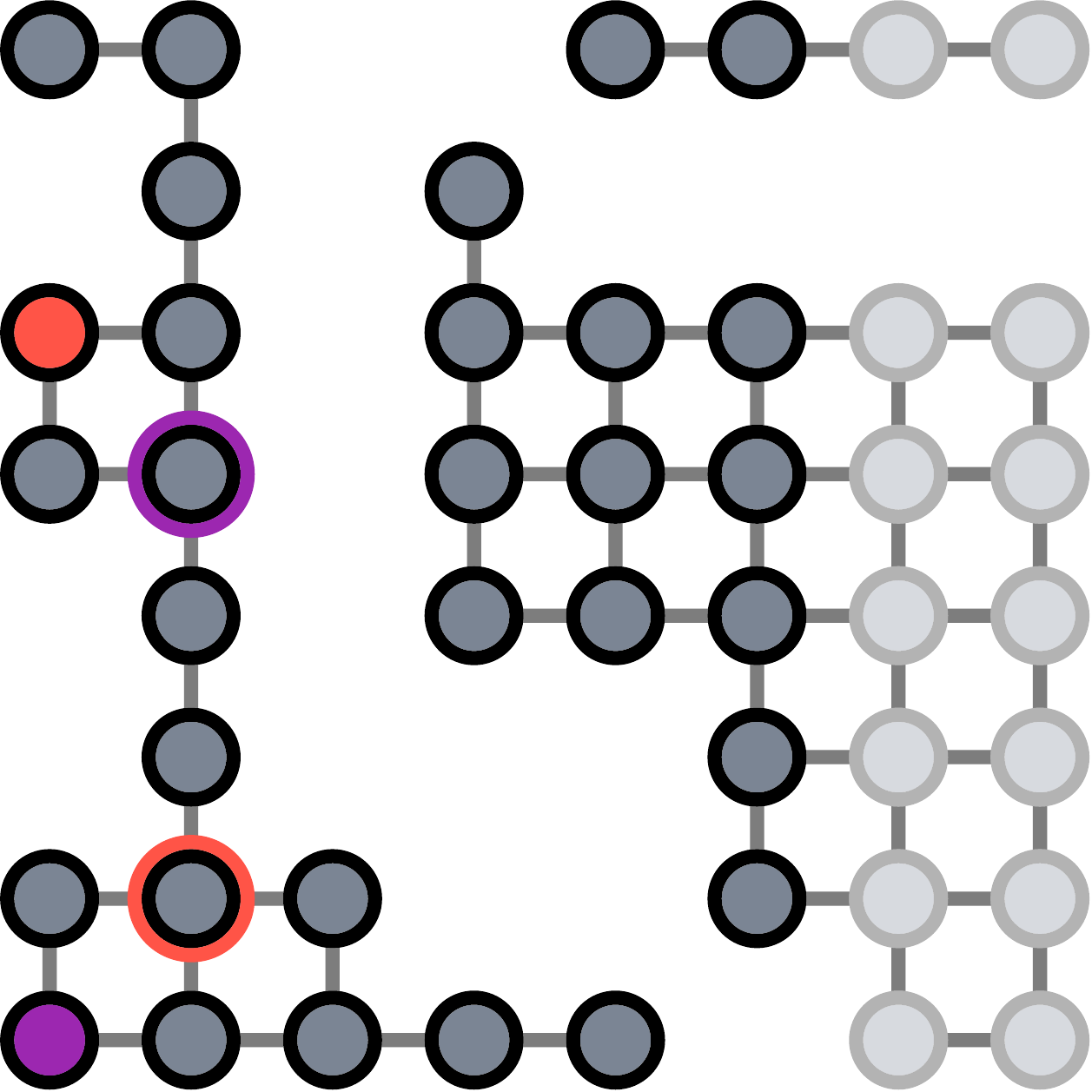}}
    \hspace{0.03\linewidth}
    \subfigure{\includegraphics[width=0.29\linewidth]{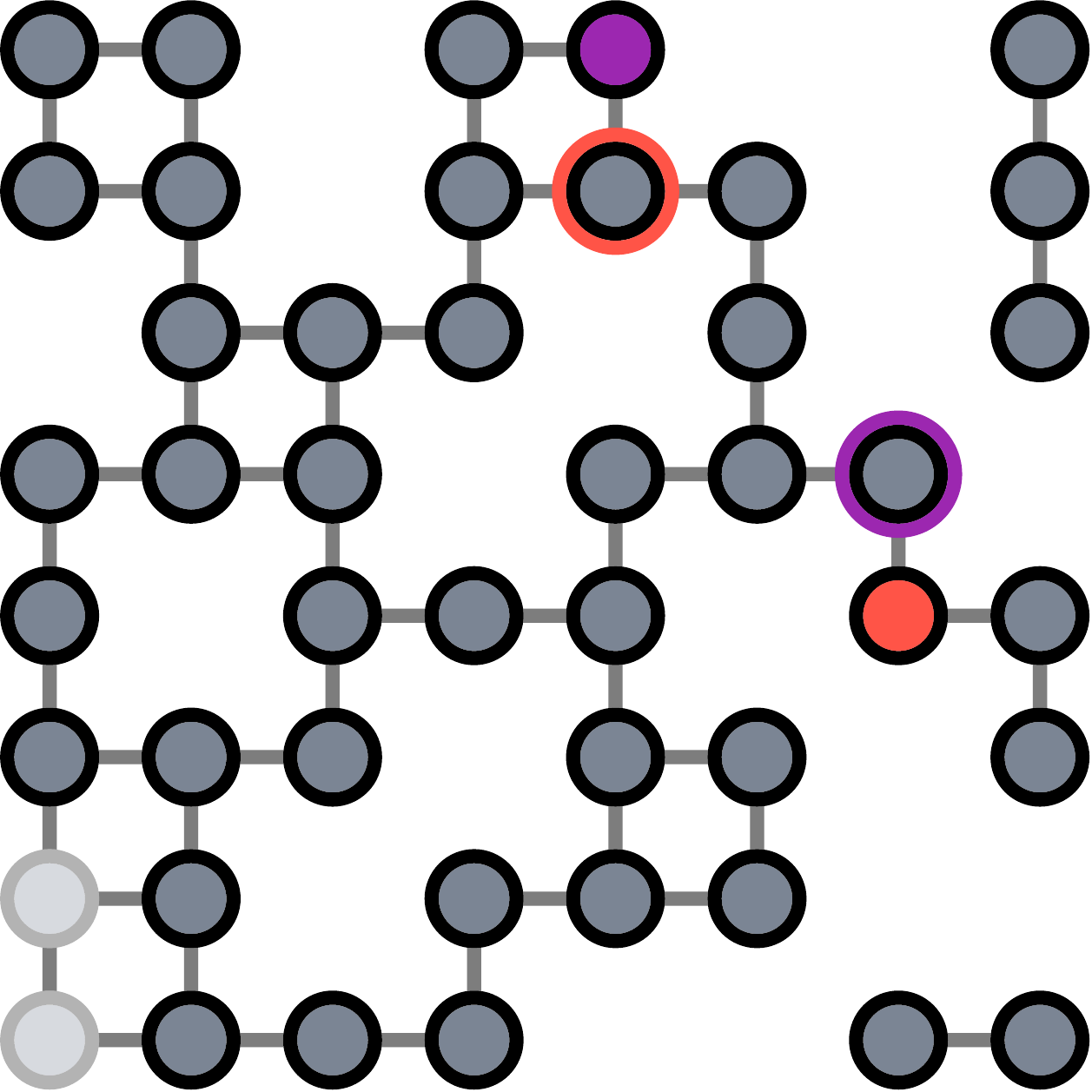}}
    \hspace{0.03\linewidth}
    \subfigure{\includegraphics[width=0.29\linewidth]{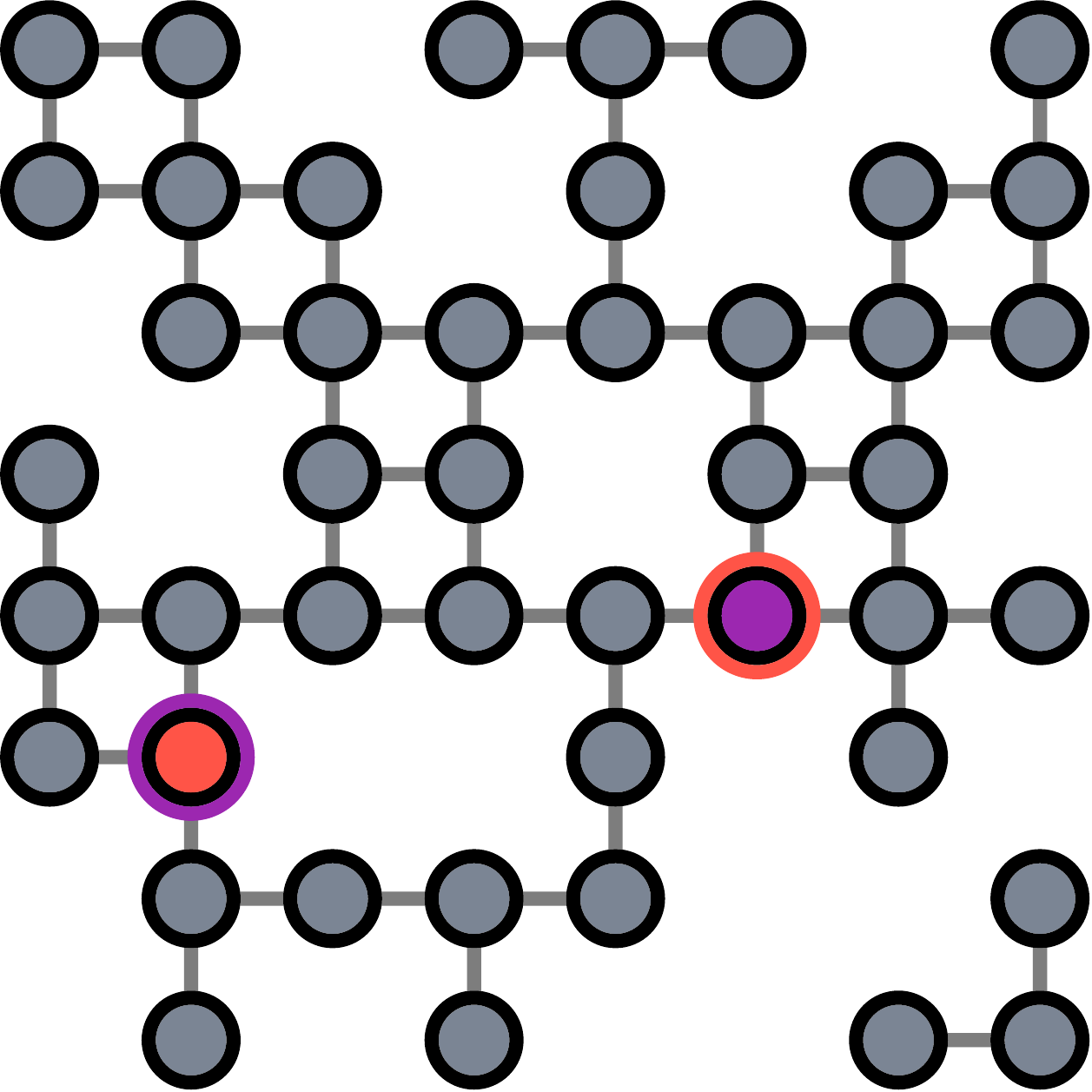}}
    \caption{Examples of complex $ 8 \times8 $ environments where agents need to use a cooperative policy to reach their goals. In all examples, the optimal paths of agents to their goals intersect, and one of them must give way to the other. Vertices that are not visible to agents are shown as transparent.} 
    \label{fig:hard70}
\end{figure*}

The results of training of the QMIX algorithm are shown in  Figure~\ref{fig:qmix-training}.  We evaluated the algorithm for every $10^5$ step on a set of unseen environments. The evaluation set was fixed throughout the training. This figure also shows evaluation curves for complex environments. 
We generated a set of complex environments so that agents needed to choose actions cooperatively, avoiding deadlocks. An example of complex environments for a environment size of $ 8 \times8 $ is shown in Figure~\ref{fig:hard70}. This series of experiments aimed to test the ability of QMIX agents to act sub-optimally for a greedy policy, but optimal for a cooperative policy.

The results of the evaluation are shown in Table~\ref{table: eval-random} for random environments and in Table~\ref{table: eval-hard} for complex environments. As a result of training, the QMIX significantly outperforms the PPO algorithm on both series of experiments, which shows the importance of using the mixing network for training.

\begin{table}[htp]
	  \caption{Comparison of the algorithms on a set of 200 environments (for each parameter set) with randomly generated obstacles. The last two columns show the success rate for PPO and QMIX, respectively. The results are averaged over three runs of each algorithm in each environment. QMIX out-performs PPO due to the use of the mixing Q-function $Q_{tot}$.}
    \label{table: eval-random}
    \centering
    \begin{tabular}{p{0.12\textwidth}p{0.12\textwidth}p{0.05\textwidth}p{0.12\textwidth}p{0.12\textwidth}p{0.12\textwidth}p{0.12\textwidth}p{0.12\textwidth}}
        \toprule
            
         $E_{size}$ & $E_{agents}$ & $R$ & $E_{horizon}$ & $E_{dist}$& $E_{density}$ & PPO & QMIX\\
        \midrule
         $8\times8$ & 2 & 5 & 16 &  5 & 0.3 & 0.539 & \textbf{0.738} \\
         $16\times16$ & 6 &5 & 32 & 6 &  0.3 & 0.614 & \textbf{0.762} \\
        $32\times32$ & 16 & 5 & 64 & 8 & 0.5 & 0.562 & \textbf{0.659} \\
        \bottomrule
    \end{tabular}
    
\end{table}

\begin{table}[htp]
	  \caption{Comparison on a set of 70 environments (for each parameter set) with complex obstacles. The last two columns show the success rate for PPO and QMIX, respectively. The results are averaged over ten runs of each algorithm in each environment. QMIX, as in the previous experiment, out-performs PPO.}
    \label{table: eval-hard}
    \centering
    \begin{tabular}{p{0.12\textwidth}p{0.12\textwidth}p{0.05\textwidth}p{0.12\textwidth}p{0.12\textwidth}p{0.12\textwidth}p{0.12\textwidth}p{0.12\textwidth}}
        \toprule
            
         $E_{size}$ & $E_{agents}$ & $R$ & $E_{horizon}$ & $E_{dist}$& $E_{density}$ & PPO & QMIX\\
        \midrule
         $8\times8$ & 2 & 5 & 16 &  5 & 0.3 & 0.454 & \textbf{0.614} \\
         $16\times16$ & 6 &5 & 32 & 6 &  0.3 & 0.541 & \textbf{0.66} \\
        $32\times32$ & 16 & 5 & 64 & 8 & 0.5 & 0.459 & \textbf{0.529} \\
        \bottomrule
    \end{tabular}
    
\end{table}

\section{Conclusion}
In this work, we considered the problem of multi-agent pathfinding in the partially observable environment. Applying state-of-the-art centralized methods that construct joint plan for all agents is not possible in this setting. Instead we rely on reinforcement learning to learn the mapping from agent's observations to actions. To learn cooperative behavior we adopted a  mixing Q-network (neural network approximator), which selects the parameters of a unifying Q-function that combines the Q-functions of the individual agents. An experimental environment was developed for launching experiments with learning algorithms. In this environment, the efficiency of the proposed method was demonstrated and its ability to outperform the state-of-the-art on-policy reinforcement learning algorithm (PPO). It should be noted that the comparison was carried out under conditions of limiting the number of episodes of interaction between the agent and the environment. If such a sample efficiency constraint is removed, the on-policy method can outperform the proposed off-policy Q-mixing network algorithm. In future work, we plan to combine the advantages of better-targeted behavior generated by the on-policy method and the ability to take into account the actions of the other agents when resolving local conflicts using QMIX. The model-based reinforcement learning approach seems promising, in which it is possible to plan and predict the behavior of other agents and objects in the environment \cite{Schrittwieser2021,Gorodetskiy2020}. We also assume that using adaptive task composition for agent training (curriculum learning) will also give a significant performance boost for tasks with a large number of agents.

\bibliography{rcai}
\bibliographystyle{acm}

\end{document}